\documentclass[
]{ceurart}

\usepackage{listings}
\usepackage{dingbat}

\usepackage{xcolor}
\definecolor{darkgreen}{rgb}{0.0, 0.5, 0.0} 

\usepackage{ccicons}
\usepackage{hyperref}
\usepackage[
    type={CC},
    modifier={by},
    version={4.0}
]{doclicense}

\lstdefinelanguage{turtle}{
    keywords={@prefix, rdf:type, rdfs:label, obo:RO_0000087},
    keywordstyle=\color{magenta}\bfseries,
    ndkeywords={rdf, rdfs, obo, cear},
    ndkeywordstyle=\color{blue}\bfseries,
    identifierstyle=\color{darkgreen}\bfseries,
    stringstyle=\color{red},
    comment=[l]{\#},
    commentstyle=\color{gray},
    morecomment=[s][\color{teal}]{<}{>},
    sensitive=true,
    morestring=[b]"
}

\lstdefinelanguage{python}{
    keywords={@prefix, rdf:type, rdfs:label, obo:RO_0000087},
    keywordstyle=\color{magenta}\bfseries,
    ndkeywords={rdf, rdfs, obo, cear},
    ndkeywordstyle=\color{blue}\bfseries,
    identifierstyle=\color{black},
    stringstyle=\color{red},
    comment=[l]{\#},
    commentstyle=\color{gray},
    morecomment=[s][\color{teal}]{<}{>},
    sensitive=true,
    morestring=[b]"
}

\begin{document}

\copyrightyear{2024}
\copyrightclause{Copyright for this paper by its authors.
  Use permitted under Creative Commons License Attribution 4.0
  International (CC BY 4.0).}

\conference{Proceedings of the Joint Ontology Workshops (JOWO) - Episode X: The Tukker Zomer of Ontology, and satellite events co-located with the 14th International Conference on Formal Ontology in Information Systems (FOIS 2024), July 15-19, 2024, Enschede, The Netherlands.}

\title{CEAR: Automatic construction of a knowledge graph of chemical entities and roles from scientific literature}
\author[1]{Stefan Langer}[%
orcid=0009-0009-9938-596X,
email=stefan1.langer@ovgu.de
]
\cormark[1]

\author[2]{Fabian Neuhaus}[%
email=fneuhaus@ovgu.de,
]

\author[3]{Andreas Nürnberger}[%
email=andreas.nuernberger@ovgu.de,
]

\address[1]{Otto-von-Guericke University Magdeburg, Universitätsplatz 2, 39106 Magdeburg, Germany}

\cortext[1]{Corresponding author.}

\begin{abstract}
    Ontologies are formal representations of knowledge in specific domains that provide a structured framework for organizing and understanding complex information. Creating ontologies, however, is a complex and time-consuming endeavor. ChEBI is a well-known ontology in the field of chemistry, which provides a comprehensive resource for defining chemical entities and their properties. However, it covers only a small fraction of the rapidly growing knowledge in chemistry and does not provide references to the scientific literature. To address this, we propose a methodology that involves augmenting existing annotated text corpora with knowledge from Chebi and fine-tuning a large language model (LLM) to recognize chemical entities and their roles in scientific text. Our experiments demonstrate the effectiveness of our approach. By combining ontological knowledge and the language understanding capabilities of LLMs, we achieve high precision and recall rates in identifying both the chemical entities and roles in scientific literature. Furthermore, we extract them from a set of 8,000 ChemRxiv articles, and apply a second LLM to create a knowledge graph (KG) of chemical entities and roles (CEAR), which provides complementary information to ChEBI, and can help to extend it.
\end{abstract}


\maketitle

\section{Introduction}
Chemistry is a large and diverse field of research with a rapidly growing number of publications available. While this is exciting and demonstrates rapid progress, the sheer volume of research texts makes it increasingly difficult to keep track of all the new discoveries and developments. Ontologies have been used to provide a structured framework for organizing this knowledge. However, manually incorporating knowledge into ontologies is a labor-intensive and time-consuming task, and therefore not feasible for all available research.

In recent years, Large Language Models (LLMs) have demonstrated exceptional performance in understanding natural language, excelling in tasks such as summarization and question answering. In this paper, we propose a novel approach that leverages the capabilities of these models to automatically create a knowledge graph (KG) of \textit{Chemical Entities And Roles} (CEAR) from research publications and to extend existing ontological knowledge.

Our approach involves automatically augmenting manually annotated text corpora with information from ChEBI, using two distinct LLMs to identify and associate chemical roles and entities, and creating a knowledge graph based on ChEBI which contains information from research texts, that is not annotated in ChEBI. We make the methodology and the resulting knowledge graph (KG) available to the research community as a basis for developing utilities to efficiently explore and structure any given set of chemistry research texts and to help with the task of extending ChEBI.

This paper is organized as follows: In Section~\ref{sec:related_work} we provide an overview of ChEBI, and methods used to create biochemical knowledge graphs and scholarly knowledge graphs, which are both relevant to our research. Section~\ref{sec:methods} outlines the steps involved in creating the KG. Here we explain our approach, providing a clear and reproducible process for others in the community to follow. In Section~\ref{sec:results}, we discuss our results for different steps in the KG creation process and the final KG. Finally, section~\ref{sec:conclusion} proposes some applications of our methods and outlines future work on this project.


\section{Related work} \label{sec:related_work}
The \href{https://www.smartprosys.ovgu.de/}{SmartProSys research initiative} aims to replace fossil raw materials in chemical production with renewable carbon sources, thus contributing to a carbon-neutral society.
The transition to sustainable and circular production processes requires research into novel chemical reaction pathways that lead from renewable raw materials via energy-efficient and low-CO2 synthesis processes to green products. The task of identifying such pathways requires the collective chemical knowledge of the world to be searched and structured in a methodical, systematic and targeted manner. This knowledge is growing rapidly: the \href{https://chemrxiv.org/engage/chemrxiv/public-dashboard}{ChemRxiv} platform, launched in 2017, already contains more than 20,000 research papers on chemistry. In addition, there are journals such as the \href{https://www.mdpi.com/journal/ijms}{International Journal of Molecular Sciences}, which has published more than 20,000 scientific articles in 2022, of which about 30-35\% are in the field of biochemistry \cite{supuran2023progress}.

\cite{chandrasekaran1999what} emphasizes that the first step in designing an effective knowledge representation system, and vocabulary, is to perform an effective ontological analysis of the field, or domain and that ontologies enable knowledge sharing.

\href{https://www.ebi.ac.uk/chebi/}{ChEBI} is a database and ontology for chemical entities of biological interest. In its November 2012 release, it contained nearly 30,000 fully annotated entities, all of which were added by expert annotators \cite{hastings2012chebi}. In 2024, ChEBI  contains almost 218,000 entities, of which more than 60,000 were fully annotated by ChEBI curators.
However, the  content of ChEBI is still very limited, when compared to data sources like \href{https://pubchem.ncbi.nlm.nih.gov/}{PubChem} with information on nearly 317 million substances and 118 million compounds\footnote{\href{https://pubchem.ncbi.nlm.nih.gov/docs/statistics}{https://pubchem.ncbi.nlm.nih.gov/docs/statistics, accessed on April 16, 2024}}.

Knowledge graphs, on the other hand, are a powerful tool for representing and querying complex, interrelated data. They are essentially a network of entities (nodes) and their interrelations (edges). The relationship between ontologies and knowledge graphs is complementary. Ontologies provide a well-defined, interconnected vocabulary, while knowledge graphs populate this vocabulary with specific real-world data instances.

Scholarly Knowledge Graphs (SKG) are structured, semantic representations of scientific data. In \cite{verma2023scholarly}, a comprehensive review is given on the field of applying machine learning, rule-based learning, and natural language processing tools and approaches to both construct SKGs and utilize them. \cite{agrawal2019scalable} uses a semi-supervised extraction approach to construct a KG from scientific text. It contains nodes of research papers with edges for citations between them. Relevant (candidate) sentences from the represented research papers are classified as \textit{aim}, \textit{method} or \textit{result} and added as nodes to the SKG. Relations connect the corresponding paper nodes to the extracted sentences, using the classified type of the sentences as type for the relations. \cite{wang2021covid} constructs knowledge graphs on COVID-19 related scientific text and creates nodes for drugs, diseases, genes and organisms. For entity extraction they use CORD-NER, a dataset with entities of the Unified Medical Language System (UMLS) annotated using distant supervision \cite{wang2020comprehensive}.

Other existing KGs are closely related to biomedical sciences.
\cite{Muhammad2020open} describes a method to construct a knowledge graph in four steps: triple extraction, triple filtering, concept linking, merging of vertices and KG population. The main principle for the triple extraction is to split the text into sentences and use a supervised open information extraction system. Triple filtering uses term frequencies to determine important concepts and remove redundant or uninformative information. The remaining concepts are annotated to clinical concepts in UMLS. The resulting KG merges vertices and links the concepts to scientific papers. \cite{rossanez2020knowledge} presents KGen, a semi-automatic method that generates KGs from scientific biomedical text using a preprocessing step that splits text into sentences, co-references and abbreviations. After a simplification process, RDF-triples are generated using part-of-speech (POS) tagging and dependency parsing. An existing model for Named Entity Recognition (NER) is used together with SPARQL to link entities to medical ontologies. The resulting KG is manually evaluated by two physicians. FORUM is a KG that links chemical entities to biomedical concepts \cite{delmas2021forum}. It is built from life-science databases and ontologies like ChEBI, ChemOnt and PubChem and uses ontological knowledge for automated reasoning and inference of relations between entities. Co-occurrence analysis in scientific literature repositories like PubMed is used to estimate the strength of the association.


\section{Methods} \label{sec:methods}
\begin{figure}
    \centering
    \includegraphics[width=1\linewidth]{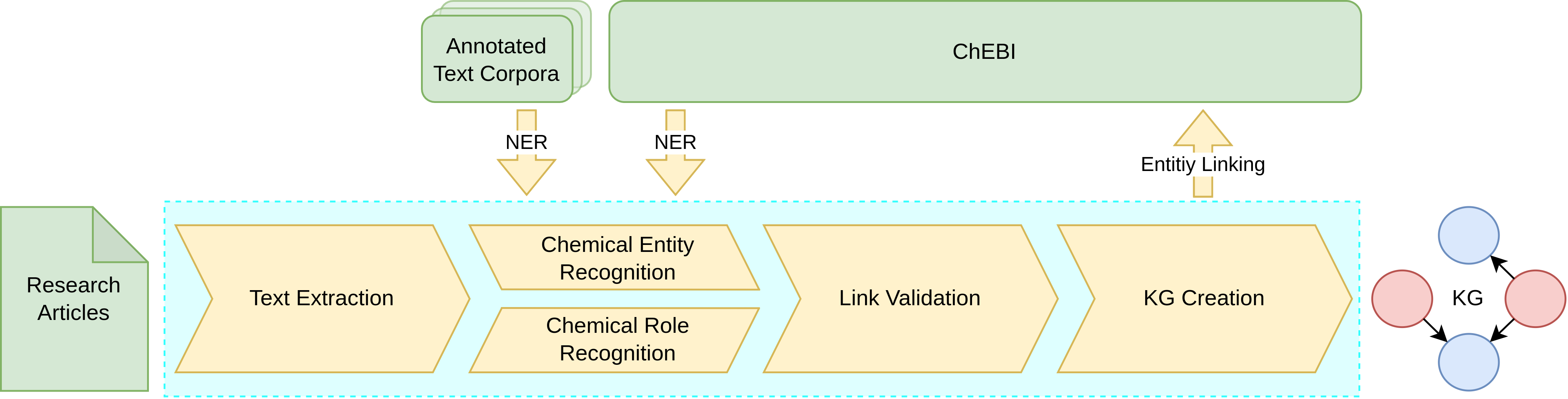}
    \caption{Working steps (yellow) and resources (green) used to create the KG (blue and red)}
    \label{fig:method}
\end{figure}
In our work, we create a KG for chemical entities and roles as defined in ChEBI. Chemical entities are atoms, substances, groups and molecules and are classified as such based on shared structural features, while roles are classified based on their activities in biological or chemical systems or their use in applications \cite{hastings2015chebi}. Figure~\ref{fig:method} outlines the method we use to create the KG: First, we extract the full text from research papers and then fine-tune an LLM to identify chemical entities and roles. Candidate sentences containing both are collected and a different LLM is used to validate the relationship between the two. Finally, we de-duplicate and normalize both chemical entities and roles, link them to the ChEBI ontology and create the KG. The following subsections explain each step in detail.

\begin{figure}
    \centering
    \includegraphics[width=0.5\linewidth]{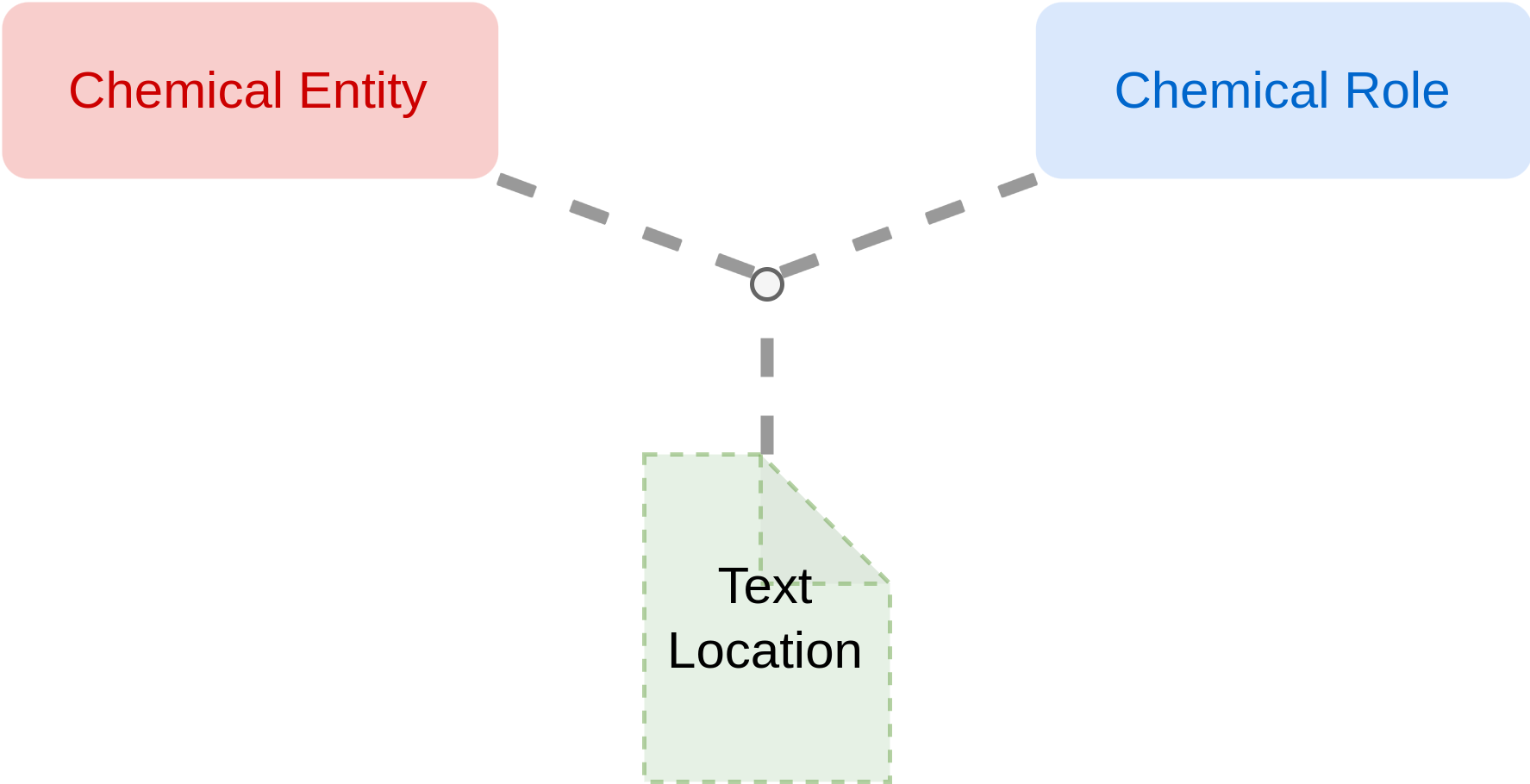}
    \caption{Information types provided by our approach.}
    \label{fig:basic-kg}
\end{figure}

Figure~\ref{fig:basic-kg} shows the different types of information provided by our approach. 
The information that is extracted from the papers has the form \lstinline|<chemical entity> has_Role <chemical Role>|, together with additional information about the text location that supports this triple.  Each text location consists of a specific paper, the page number in the paper, and the character position of the sentence relative to that page number. RDF is not ideal to model these relations because it does not allow to annotate a triple with its source without clumsy workarounds (e.g., reification of triples). Thus, we plan to release a KG built using RDF-star. The current RDF version does not include any text locations.

\subsection{Text extraction from research papers}\label{sec:methods-extract}
Research papers are a rich source of information. They contain author names, images, tables, citations, bibliographies, and more. To address the challenge of extracting the papers' full texts in an efficient way, we chose a very simple approach which involves using a Linux utility called \lstinline|pdftotext|. While it cannot identify floating objects in plain text, such as image and table captions or footers and page numbers, it can reliably extract different formats, ranging from one-column to two-column styles.

We downloaded a set of 8,000 chemistry research papers from various categories of \href{https://chemrxiv.org/}{ChemRxiv} and extracted their full text as JSON documents, including information about the page it was extracted from. Content-based checksums ensure that no duplicates are processed, even when crawling other sources for research papers. The checksums are also used as identifiers between the original PDF file and the associated JSON document.

\subsection{Chemical entity and role recognition}\label{sec:methods-ner}
Transformer-based Large Language Models (LLMs) have proven effective in understanding language patterns and thus in Natural Language Processing (NLP) tasks such as \textit{Named-Entity-Recognition} (NER), which we use in order to identify chemical entities and roles. Approaches such as RoBERTa or BERT use \textit{masked language modeling} (MLM), where some tokens in an input sequence are randomly masked and the model is trained to predict the original token \cite{devlin2019bert}. Electra models use a pre-training task called \textit{replaced token detection} or \textit{token discrimination}, where instead of predicting a masked token, a discriminative model is trained to predict whether a token in the corrupted input sequence was replaced by a generator sample. We chose this approach because it is more sample-efficient \cite{clark2020electra}, and fine-tuned a \href{https://huggingface.co/google/electra-base-discriminator}{pre-trained Electra model} on three different datasets:
\begin{itemize}
    \item The \textit{\href{https://github.com/JHnlp/BioCreative-V-CDR-Corpus}{BC5CDR}} dataset consists of human annotations of chemicals, diseases and their interactions from 1,500 PubMed articles \cite{wei2016assessing}.
    \item The \emph{\href{https://ftp.ncbi.nlm.nih.gov/pub/lu/NLMChem/}{NLM-Chem corpus}} contains 150 full-text articles on biomedical literature, carefully selected for containing chemical entities which are difficult to find for NER tools. Ten domain experts annotated the chemical entities in three annotation rounds \cite{rezarta2021nlm}.
    \item The \emph{\href{https://github.com/UCDenver-ccp/CRAFT}{CRAFT corpus}} contains 97 full-text open access articles from the \href{https://www.ncbi.nlm.nih.gov/pmc/tools/openftlist/}{PubMed Central Open Access subset}. It identifies all mentions of nearly all concepts from nine prominent biomedical ontologies, including ChEBI \cite{cohen2017colorado}.
\end{itemize}
A fourth manually annotated dataset, \textit{\href{https://ftp.expasy.org/databases/rhea/nlp/}{EnzChemRED}} \cite{lai2024enzchemred}, provides chemical entities and proteins, as well as conversions during chemical reactions. It is highly relevant to our NER task, but given its recent availability, it has not yet been used for fine-tuning.

The CRAFT corpus annotates all entities according to nine different ontologies from different areas of interest. Chemical annotations, including chemical entities and roles are provided along an extension of an older version of the ChEBI ontology. Although the NLM-Chem corpus and the BC5CDR dataset also annotate all chemicals in the provided full texts, and although BC5CDR annotates diseases, they do not include any chemical roles, such as ligand, acid, buffer, or catalyst. To overcome this limitation, we used a semi-supervised approach and automatically annotated all roles defined by their label and synonyms in ChEBI using a lexical approach. We ignore all role strings that are shorter than four characters to avoid mislabeling identical strings with different meanings (homonyms).

\subsection{Link Validation}\label{sec:methods-linking}
\begin{figure}
    \centering
    \includegraphics[width=1\linewidth]{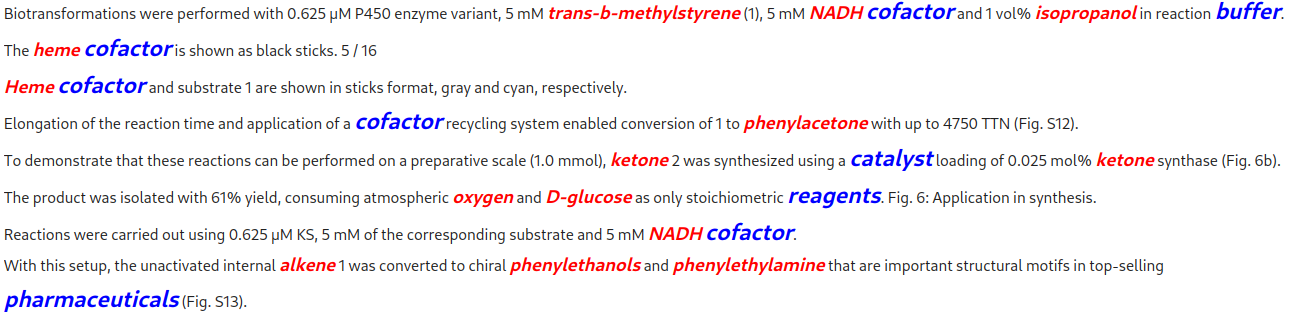}
    \caption{Sample sentences with inferred chemical entities (red) and roles (blue).}
    \label{fig:role-infer}
\end{figure}

We applied the fine-tuned Electra model to the extracted text of the chemistry research papers, collecting all sentences, that contained at least one chemical entity and at least one chemical role (Figure \ref{fig:role-infer}). For each sentence, we store the exact text location and the inferred chemical entities and roles.

The co-occurrence of chemical entities and roles within the same text block suggests that the chemical entity may have this specific role. However, this correlation alone is not sufficient to draw a definitive conclusion. To address this, we use another LLM to verify the role of a chemical entity based on the given contextual information.
\emph{LLAMA~2} is a collection of pre-trained and fine-tuned LLMs ranging in size from 7 billion to 70 billion parameters. \emph{LLAMA~2-CHAT} is specifically trained for conversational tasks using reinforcement learning with human feedback (RLHF) \cite{touvron2023llama}.

\begin{figure}
    \centering
    \includegraphics[width=1\linewidth]{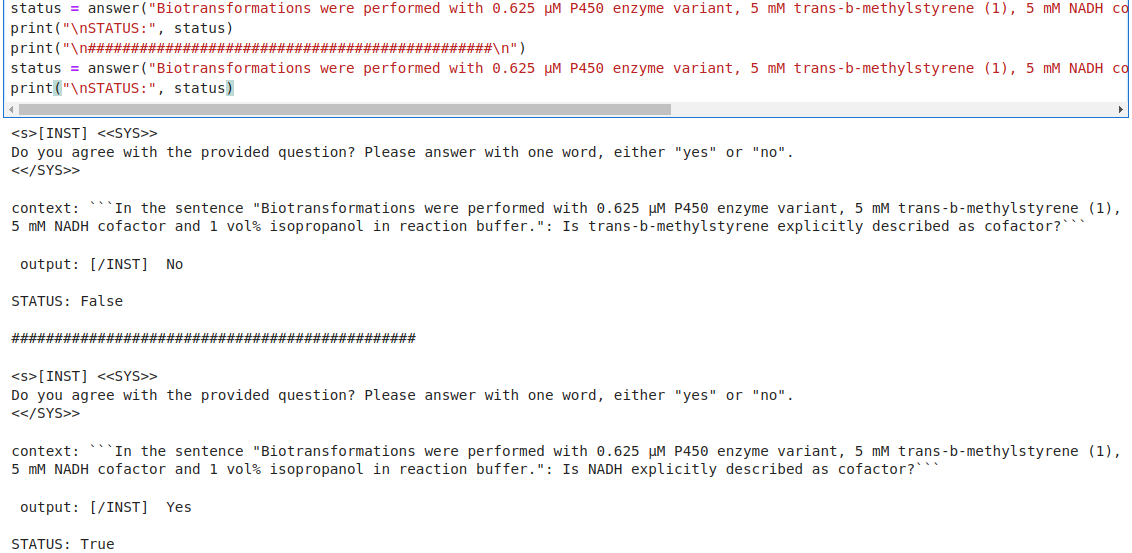}
    \caption{Question answering using LLAMA-2-CHAT}
    \label{fig:llama2}
\end{figure}

In this paper we used \href{https://huggingface.co/meta-llama/Llama-2-7b-chat-hf}{LLAMA-2-7b-CHAT} and split the prompt into:
\begin{itemize}
    \item a \textit{system prompt}, that defines the role of the LLM and makes sure that it simply confirms or rejects the relation between chemical entity and chemical role without any further explanations or other context that could complicate the parsing of the answer. For this paper we used:
    \begin{lstlisting}[language=python, basicstyle=\footnotesize\ttfamily]
system_prompt = "Do you agree with the provided question? Please answer with one
    word, either 'yes' or 'no'."
    \end{lstlisting}
    \item a \textit{user prompt}, that presents the context to the LLM along with the question whether, according to the given context, a specific chemical entity has a specific role. For this paper we used:
    \begin{lstlisting}[language=python, basicstyle=\footnotesize\ttfamily]
user_prompt = f"In the sentence '{sentence}': Is {chemical} explicitly described
    as {role}?"
    \end{lstlisting}
\end{itemize}
A \lstinline|temperature| hyperparameter of 0.1 and a \lstinline|top-p| of 0.95 ensure a somewhat deterministic behavior and reproducible results. All confirmed relations, as well as the associated \textit{sentence location}, the \textit{chemical entity}, and the \textit{role}, are collected for the construction of the KG, while the remaining discarded relations are stored for analysis.
Figure~\ref{fig:llama2} shows how LLAMA-2 answers the questions whether \textit{trans-b-methylstyrene} or \textit{NAOH} is described as \textit{cofactor} in the given sentence (see the first sentence in Figure~\ref{fig:role-infer} for a visualization of the same sentence with its chemical entities and roles rendered in red and blue).

\subsection{Knowledge graph creation}\label{sec:methods-kg}
From the confirmed relationships, we group all chemical entities and roles using the labels and synonyms from ChEBI. If an entity is not part of ChEBI, we use its original appearance in the text. For this we only use chemical entities and roles with a character length of at least 2. For each pair of chemical entity and role, we count how many references to specific text locations exist. A higher frequency of occurrence of a relation increases our confidence in both, its correct identification in the research text and the correctness of its meaning. At the same time, it also reduces the novelty of the identified information. A hyperparameter \lstinline|minRef|, which simply ignores relations with a low frequency, can be used to increase precision at the expense of recall or vice versa.

The knowledge graph consists of the described relations. It is stored using the \textit{Terse RDF Triple Language} (Turtle). All chemical entities (\lstinline[language=turtle]|obo:CHEBI_24431|) and roles (\lstinline[language=turtle]|obo:CHEBI_50906|) that are known to ChEBI are defined using their ChEBI identifier.

Chemical entities or roles that are unknown to ChEBI are defined using the \lstinline[language=turtle]|@prefix cear: <https://wwwiti.cs.uni-magdeburg.de/iti_dke/cear/> .| namespace. The \lstinline[language=turtle]|obo:RO_0000087| is used in ChEBI to define roles of chemical entities.

The following listing shows an example for two chemical entities, \textit{ethylene glycol bis(2-aminoethyl)tetraacetate} and \textit{PBS}, both of which have the role \textit{buffer}:

\begin{lstlisting}[language=Turtle, basicstyle=\footnotesize\ttfamily]
@prefix rdf: <http://www.w3.org/1999/02/22-rdf-syntax-ns#> .
@prefix rdfs: <http://www.w3.org/2000/01/rdf-schema#> .
@prefix obo: <http://purl.obolibrary.org/obo/> .
@prefix cear: <https://wwwiti.cs.uni-magdeburg.de/iti_dke/cear/> .

obo:CHEBI_35225 rdf:type obo:CHEBI_50906 .
obo:CHEBI_35225 rdfs:label "buffer" .

obo:CHEBI_30741 rdf:type obo:CHEBI_24431 .
obo:CHEBI_30741 rdfs:label "ethylene glycol bis(2-aminoethyl)tetraacetate" .
obo:CHEBI_30741 obo:RO_0000087 obo:CHEBI_35225 .

cear:chem_4023 rdf:type obo:CHEBI_24431 .
cear:chem_4023 rdfs:label "PBS" .
cear:chem_4023 obo:RO_0000087 obo:CHEBI_35225 .
\end{lstlisting}

\section{Results} \label{sec:results}

\subsection{Chemical entity and role recognition}
\begin{table*}
    \centering
    \caption{Results for different combinations of fine-tune and evalutaion corpora (strict spans), \cite{rezarta2021nlm} results in italic}
    \label{tab:all-train-eval}
    \begin{tabular}{lcccccccccc}
        \toprule
        \textbf{Train Corpus} & \textbf{Type} & \multicolumn{3}{l}{\textbf{Eval on BC5CDR}} & \multicolumn{3}{l}{\textbf{Eval on NLM-Chem}} & \multicolumn{3}{l}{\textbf{Eval on CRAFT}}\\
        \cmidrule(lr){3-5} \cmidrule{6-8} \cmidrule{9-11}
        & & \textbf{P} & \textbf{R} & \textbf{F1}
        & \textbf{P} & \textbf{R} & \textbf{F1}
        & \textbf{P} & \textbf{R} & \textbf{F1} \\
        \toprule
        BC5CDR & chem & 94.2 & 90.6 & 92.4 & 75.9 & \underline{54.3} & \underline{63.3} & \underline{63.3} & \underline{30.4} & \underline{41.1} \\
        BC5CDR & role & 89.5 & 90.7 & 90.1 & 84.7 & 83.1 & 83.9 & 75.4 & 59.1 & 66.3 \\
        \midrule
        NLM-Chem & chem & 90.3 & 80.8 & 85.3 & 85.8 & 76.8 & 81.1 & \underline{68.0} & \underline{40.2} & \underline{50.5} \\
        NLM-Chem & role & 70.2 & 82.2 & 75.7 & 83.1 & 89.7 & 86.3 & 79.5 & 76.2 & 77.8 \\
        \midrule
        CRAFT & chem & 85.3 & \underline{67.2} & 75.2 & \underline{65.4} & \underline{44.8} & \underline{53.2} & 93.4 & 85.1 & 89.0 \\
        CRAFT & role & 65.4 & 63.6 & 64.5 & 81.4 & 77.9 & 79.6 & 93.6 & 92.6 & 93.1 \\
        \midrule
        \textit{NLM+BC5CDR \cite{rezarta2021nlm}} & \textit{chem} & - & - & - & \textit{81.0} & \textit{71.1 } & \textit{75.7} & - & - & - \\
        NLM+BC5CDR & chem & 93.4 & 90.2 & 91.8 & \textbf{85.2} & \textbf{77.5} & \textbf{81.2} & 68.1 & \underline{39.9} & \underline{50.3} \\
        NLM+BC5CDR & role & 91.5 & 92.0 & 91.7 & 92.3 & 93.9 & 93.1 & 79.5 & 76.2 & 77.8 \\
        \midrule
        NLM+CRAFT & chem & 90.4 & 78.3 & 83.9 & 84.0 & 70.9 & 76.9 & 88.0 & 74.1 & 80.4 \\
        NLM+CRAFT & role & 79.0 & 83.4 & 81.1 & 88.5 & 92.1 & 90.2 & 87.1 & 90.3 & 88.7 \\
        \midrule
        all corpora & chem & 92.0 & 89.2 & 90.6 & 84.4 & 71.2 & 77.3 & 89.2 & 74.0 & 80.9 \\
        all corpora & role & 89.8 & 91.6 & 90.7 & 90.5 & 93.7 & 92.1 & 87.3 & 92.2 & 89.7 \\
        \bottomrule
    \end{tabular}
\end{table*}

In section~\ref{sec:methods-ner}, we discussed how we used the BC5CDR corpus, the NLM-Chem corpus and the CRAFT corpus to fine-tune our Electra model for NER. As in \cite{rezarta2021nlm}, we counted a prediction as a true positive only if both the start and end locations of the characters of the complete entity exactly matched. This is a very strict definition, since the complexity of chemical entities makes it difficult to identify exact boundaries of entities or word tokens, for example: \textit{dipotassium 2-alkylbenzotriazolyl bis(trifuoroborate)s, 4,7-dibromo-2-octyl-2,1,3-benzotriazole}\cite{rezarta2021nlm}.

Table~\ref{tab:all-train-eval} shows the precision, recall and f-measure when fine-tuned on only one or multiple of the corpora. We have included cross-corpus evaluation data, and we can see that a model fine-tuned on the NLM-Chem or BC5CDR corpus performs very poorly when evaluated on the CRAFT corpus. Similarly, when a model fine-tuned using CRAFT is evaluated on NLM-Chem, the results are very poor. This indicates a lack of generalizability across datasets. Table~\ref{tab:eval-details} shows the reason by listing the ten most frequent misclassifications. All of the text corpora were manually curated to annotate all chemical entities contained in the texts. However, despite their common goal, they show discrepancies in annotation. For example, the chemical entities "DNA", "RNA" and "mRNA" are annotated in the CRAFT corpus, but not in the NLM-Chem corpus, hence the false negatives. The character "b", that appears as a false positive when a model fine-tuned on NLM-Chem is evaluated on CRAFT, is used in genetics to describe base pairs of DNA or RNA. Similarly, "PBS" is marked as a chemical entity in the NLM-chem corpus, but in CRAFT it is neglected. This illustrates how, depending on the context or background of the annotators, or depending on their research goals, there may be disagreement about which entities are considered chemical and which are not. While, for instance, a person working in criminal forensics might treat DNA as a chemical and focus on ways to identify it in a given substance, a biologist might treat DNA more as a biological concept. Therefore, the poor out-of-distribution performance is unavoidable: Even if a model could be created that perfectly aligns with the opinions of the NLM-Chem experts, the CRAFT experts might not agree.

\cite{rezarta2021nlm} reports a precision of 81.0~\%, a recall of 71.1~\% and an F1-measure of 75.7~\% in identifying chemical entities when fine-tuned on both the NLM-Chem and the BC5CDR corpus and evaluated on NLM-Chem using Bluebert (italic results in the table). Our results demonstrate a better precision of 85.2~\%, a better recall of 77.5~\%, and consequently, a better F1-measure of 81.2~\% (bold results in the table). However, when all corpora are employed for fine-tuning the LLM, the recall rate drops to 71.2~\%. We attribute this deterioration to the described disagreement between different groups of annotators. Since we want to provide a comprehensive understanding of chemical entities and their roles in our KG, we still use this model for the subsequent steps.

Since we only lexically annotated roles from ChEBI in the NLM-Chem corpus with a minimum length of 4 characters (see section~\ref{sec:methods-ner}), "dye" is one of the most common false positive roles when evaluating a model fine-tuned with CRAFT on the NLM-Chem corpus.\footnote{Experiments with a minimum length of 3 characters led to a large drop in both precision and recall when evaluated on the CRAFT corpus.} From the results we can still see high precision and recall rates for roles, when a model that is fine-tuned on NLM-Chem and BC5CDR is evaluated on CRAFT. The same applies to models fine-tuned using all corpora. This demonstrates, that the described semi-supervised lexical approach is effective.

\begin{table*}
    \centering
    \caption{Most frequently misclassified entities for cross-dataset-evaluation}
    \label{tab:eval-details}
    \begin{tabular}{llcccc}
        \toprule
        \textbf{Fine-Tuned on} & \textbf{Eval on} &
        \multicolumn{2}{c}{\textbf{False Positives}} &
        \multicolumn{2}{c}{\textbf{False Negatives}} \\
        \cmidrule(lr){3-4} \cmidrule(lr){5-6}
        & & \textbf{String} & \textbf{count} & \textbf{String} & \textbf{count} \\
        \toprule
        \multicolumn{6}{l}{\textbf{Misclassified chemical entities}} \\
        \midrule
        \multirow{10}{*}{NLM-Chem} &
        \multirow{10}{*}{CRAFT}
          & PBS & 68 & protein & 285 \\
        & & huntingtin & 26 & DNA & 113 \\
        & & tet & 16 & proteins & 108 \\
        & & polyglutamine & 12 & A$\beta$ & 106 \\
        & & paraffin & 9 & b & 79 \\
        & & Alcian blue & 8 & RNA & 64 \\
        & & pachytene & 7 & mRNA & 47 \\
        & & Alexa & 3 & solution & 41 \\
        \midrule
        \multirow{10}{*}{CRAFT} &
        \multirow{10}{*}{NLM-Chem}
        & protein & 116 & EGCG & 139 \\
        & & glucose & 77 & BAK & 128 \\
        & & solution & 76 & DEX-IND & 125 \\
        & & GDP & 57 & CKC & 118 \\
        & & proteins & 56 & PTX & 114 \\
        & & DAT & 39 & CDA & 102 \\
        & & mixture & 37 & VAM3 & 95 \\
        & & ADP & 36 & AEATP & 94 \\
        \toprule
        \multicolumn{6}{l}{\textbf{Misclassified role entities}} \\
        \midrule
        \multirow{10}{*}{NLM-Chem} &
        \multirow{10}{*}{CRAFT}
          & acid & 15 & dye & 10 \\
        & & rogen & 5 & chow & 10 \\
        & & agonist & 4 & androgen & 4 \\
        & & inhibitors & 3 & acidic & 4 \\
        & & activator & 2 & pigment & 4 \\
        & & acids & 2 & pigmented & 4 \\
        & & BMP inhibitor & 1 & PPAR$\rho$ agonist & 4 \\
        & & acceptor & 1 & epitopes & 3 \\
        \midrule
        \multirow{10}{*}{CRAFT} &
        \multirow{10}{*}{NLM-Chem}
        & epitopes & 22 & donor & 33 \\
        & & biomarkers & 14 & catalyst & 13 \\
        & & biocides & 10 & agonist & 12 \\
        & & inhibitors & 8 & base & 10 \\
        & & buffer & 6 & acceptor & 10 \\
        & & inhibitor & 5 & agonists & 7 \\
        & & pharmacological & 5 & antidiabetic & 7 \\
        & & hormone & 4 & carrier & 7 \\
        \bottomrule
    \end{tabular}
\end{table*}

\begin{figure}
    \centering
    \includegraphics[width=1\linewidth]{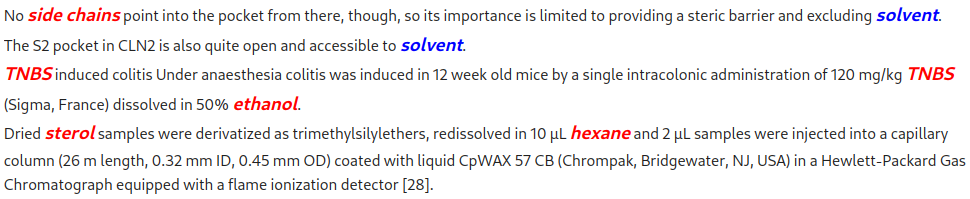}
    \caption{While \textit{solvent} is annotated in CRAFT (blue), \textit{dissolved} and \textit{redissolved} are not.}
    \label{fig:only-role-nouns}
\end{figure}

In CRAFT, chemical roles are annotated only if they appear as nouns, but not, if they are paraphrased with other words. Similarly, our lexical approach for both the BC5CDR and the NLM-Chem corpus considers only nouns. Figure~\ref{fig:only-role-nouns} shows some manually annotated text from the CRAFT corpus, with chemical roles rendered in blue and chemical entities in red. It shows that \textit{solvent} is annotated as a role, while \textit{dissolved} and \textit{redissolved} are not. While this may be correct from an annotator's point of view, it limits the expressiveness of the current version of our KG.

\subsection{Link Validation and Knowledge Graph Construction}
\begin{figure}
    \centering
    \includegraphics[width=0.85\linewidth]{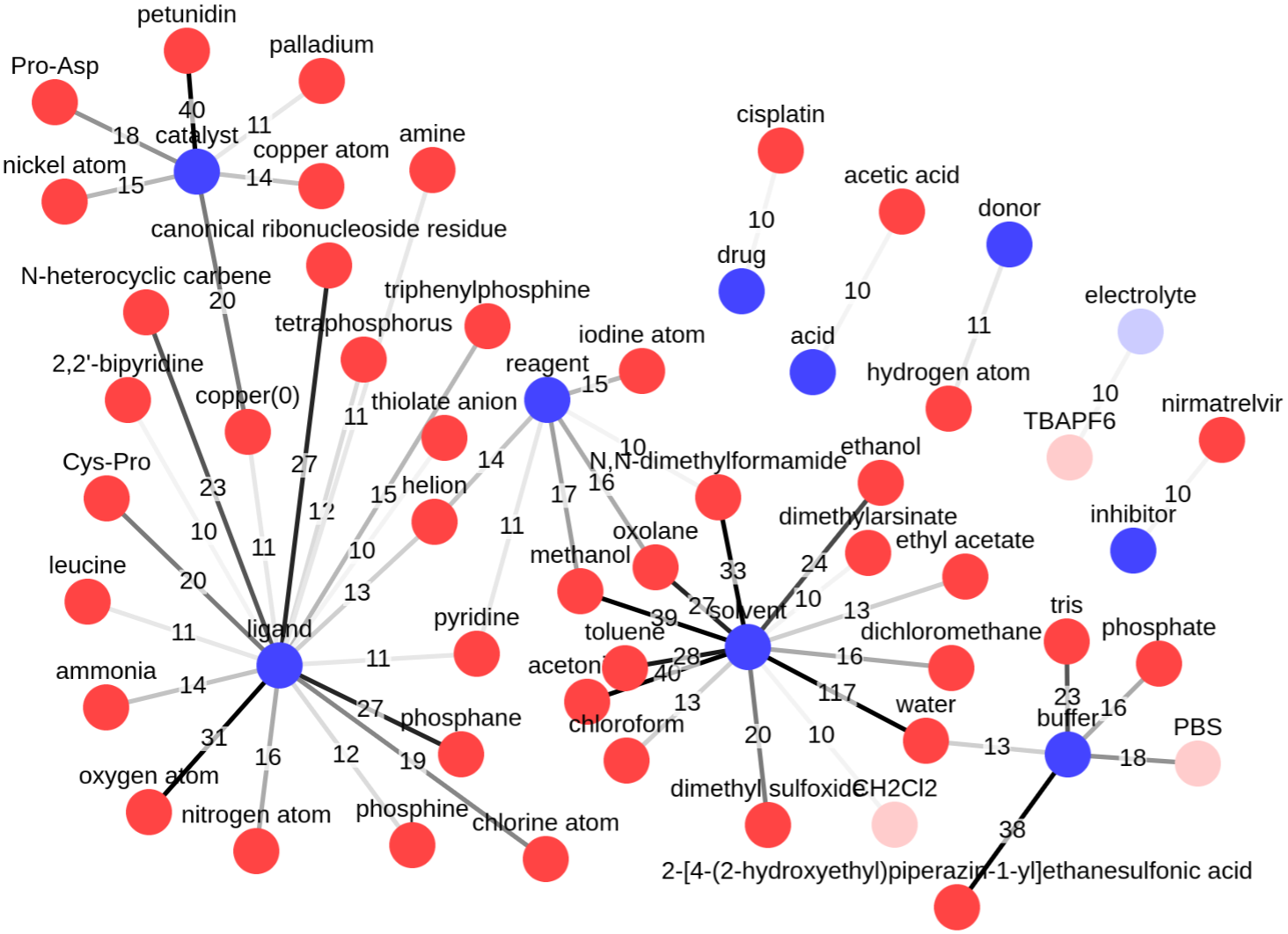}
    \caption{Chemical entities (red) and roles (blue) on a small set of papers using \lstinline|minRef=10|}
    \label{fig:kg_small}
\end{figure}

After applying the LLAMA-2 model for the validation of links between chemical entities and roles, and after grouping and applying the \lstinline|minRef| hyperparameter as discussed in section~\ref{sec:methods-kg}, two representations of the resulting KG are available. An RDF representation and a graph representation for HTML that shows chemical entities and roles as nodes, and the \lstinline|has_role| relation as an edge connecting these nodes. Figure~\ref{fig:kg_small} shows a sample graph generated on a small subset of the actual 8,000 papers, with a \lstinline|minRef| hyperparameter of 10. The dark red nodes represent chemical entities available in ChEBI, while the light red nodes represent additional chemical entities unknown to ChEBI. Similartly, the dark blue nodes represent chemical roles available in ChEBI, and the light blue nodes represent other chemical roles. The edges are labeled with the frequency with which a given relation is mentioned in the literature set. To improve the visual clarity of the graph, we have adjusted the colors of the edges based on these numbers. The darker an edge appears, the stronger the relation between the chemical entity and the role in our literature. Please be aware that due to the settings for \lstinline|minRef|, all relations with a frequency lower than 10 are ignored. Consequently, this graph shows only a very limited number of very common chemical entities with their roles in a small set of research papers.

To determine associations between chemical entities and roles, we applied the LLAMA-2 model to 115,537 candidate sentences, that contained at least one chemical entity and one role. During this step, 58,511 relations were confirmed and 272,053 were rejected. The number of candidate sentences is not the sum of confirmed and rejected relations, because each sentence can have multiple chemical entities and roles and we check all combinations.

\begin{table*}
    \centering
    \caption{Most and least frequent relations in KG}
    \label{tab:kg_relation_freq}
    \begin{tabular}{llllc}
        \toprule
        \textbf{source} & \textbf{chemical entity} & \textbf{source} & \textbf{chemical role} & \textbf{count} \\
        \midrule
        ChEBI & water & ChEBI & solvent & 1,085 \\
        ChEBI & methanol & ChEBI & solvent & 551 \\
        ChEBI & dimethyl sulfoxide & ChEBI & solvent & 438 \\
        ChEBI & N,N-dimethylformamide & ChEBI & solvent & 402 \\
        ChEBI & oxolane & ChEBI & solvent & 398 \\
        ChEBI & acetonitrile & ChEBI & solvent & 388 \\
        ChEBI & 2-[4-(2-hydroxyethyl)piperazin-1-yl]e... & ChEBI & buffer & 375 \\
        ChEBI & tris & ChEBI & buffer & 271 \\
        ChEBI & ethanol & ChEBI & solvent & 268 \\
        ChEBI & toluene & ChEBI & solvent & 268 \\
        \textbf{CEAR} & PBS & ChEBI & buffer & 249 \\
        \midrule
        \textbf{CEAR} & 1-propionyl-d-lysergic acid diethylam... & ChEBI & drug & 1 \\
        \textbf{CEAR} & tetracetate & ChEBI  & ligand & 1 \\
        \textbf{CEAR} & peroxysulfate(2-) & ChEBI & oxidising agent & 1 \\
        \textbf{CEAR} & 2-[4-(2-hydroxyethyl)piperazin-1-yl]et... & \textbf{CEAR} & buffers & 1 \\
        ChEBI & 5-fluorouracil & ChEBI & antineoplastic agent & 1 \\
        \textbf{CEAR} & SiCl4 + 4SO2 + 4MeCl (10) Thionyl chl... & ChEBI & reagent & 1 \\
        ChEBI & phenylacetonitrile & ChEBI & nucleophilic agent & 1 \\
        \textbf{CEAR} & $\alpha$-chloroamide & ChEBI & cofactor & 1 \\
        \textbf{CEAR} & Cu–t-Bu-BDPP & \textbf{ChEBI} & catalyst & 1 \\
        \bottomrule
    \end{tabular}
\end{table*}
Table~\ref{tab:kg_relation_freq} shows the most and the least frequent relations between chemical entities and roles in our set of texts. For example, water was described as a solvent in 1,085 sentences out of our 8,000 research papers. We can see, that almost all of the chemical entities and roles of the top relations are already annotated in ChEBI. The least frequent relations mostly show CEAR chemical entities (which are unknown to ChEBI). For better visibility we have marked them in bold. Please note that we could not group CEAR entities, because we do not know about their synonyms. This fact leads to an under-representation of CEAR chemical entities and roles in the high-frequency relations of our results. Please also note, that the role "buffers" was not identified as a ChEBI role: While some roles, such as "solvent" or "ligand" are annotated with their plural forms as a synonym in ChEBI, "buffer" is not.

Table~\ref{tab:kg_stats} shows some information about the KG, when created using different settings for \lstinline|minRef|\footnote{All versions of the KG can be downloaded at: \href{https://wwwiti.cs.uni-magdeburg.de/iti_dke/cear/}{https://wwwiti.cs.uni-magdeburg.de/iti\_dke/cear/}.}. We can see that if we increase the \lstinline|minRef| hyperparameter to only 2, the number of relations, relevant text positions, distinct chemical entities and roles decreases drastically. Raising \lstinline|minRef| effectively trades recall and novelty for a better precision and  a higher rate of well-known facts.

\begin{table*}
    \centering
    \caption{KG statistics for different settings of minRef}
    \label{tab:kg_stats}
    \begin{tabular}{lcccccc}
        \toprule
         & \multicolumn{6}{c}{\textbf{minRef settings}}\\
        \cmidrule(lr){2-7}
        & \textbf{1} & \textbf{2} & \textbf{5}
        & \textbf{10} & \textbf{20} & \textbf{50} \\
        \midrule
        number of relations & 28,038 & 6,586 & 1,488 & 547 & 232 & 60 \\
        number of relevant text positions & 57,846 & 36,394 & 23,999 & 18,088 & 13,932 & 9,049 \\
        \midrule
        distinct chemical entities (ChEBI) & 3,680 & 1,813 & 686 & 300 & 158 & 50 \\
        distinct chemical entities (CEAR) & 13,818 & 2,210 & 233 & 63 & 17 & 4 \\
        \midrule
        distinct chemical roles (ChEBI) & 214 & 126 & 69 & 37 & 25 & 11 \\
        distinct chemical roles (CEAR) & 455 & 75 & 7 & 3 & 1 & 0 \\
        \bottomrule
    \end{tabular}
\end{table*}

The prompts used to confirm or reject relationships using LLAMA-2 also have a big impact on the results. After modifying the system prompt slightly from asking for "\lstinline|one word, either 'yes' or 'no'|" as an answer to asking for only "\lstinline|one word|", we had 12 times fewer chemical entities and roles and 2.3 times fewer relationships between them. Adding additional text to the system prompt, such as "\lstinline|You are an expert in chemistry|", sometimes changed the answer to include long explanations about why the answer was "yes" or "no". Changing the user prompt to consider only information described in the sentence, which is what we want when constructing a KG from research papers, resulted in 2.3 times fewer confirmed relations and 2.1 times fewer chemical entities and roles. For this paper we decided to use very restrictive questions in the hope for a KG with a higher precision.

In order to evaluate the overall quality of the constructed KG, three methods can be used: Gold standard-based evaluation, manual evaluation with domain experts and annotators, and application-based evaluation with competency questions \cite{verma2023scholarly}. The latter involves asking questions and answering them using the constructed KG.

We are currently assessing the two following ideas:
 \begin{itemize}
     \item \textit{Automatic evaluation using gold standards}: An existing KG or ontology can be used as a gold standard and applying automated reasoning. However, to the best of our knowledge, there are no gold standards in literature for evaluating triples extracted from unstructured texts about chemical entities and their roles. Even for existing chemical entities and roles in ChEBI, the relations between them are not fully annotated. We are currently researching, whether we can use a combination of ChEBI and PubChem or other databases to get meaningful evaluation results.
     \item \textit{Manual evaluation with domain experts}: Precision can be determined by letting experts evaluate the rejected and confirmed relations between chemical entities and their roles in the collected sentences. To determine recall of the final KG, experts would need to manually annotate all relations between chemical entities and their roles in a fixed set of scientific texts. This task is not trivial and involves decisions such as, whether to consider only nouns (like in the CRAFT corpus) or also verbs describing a specific role (e.~g.: "dissolved" for "solvent"), or whether to use intrinsic knowledge about chemical entities.
 \end{itemize}
Although the resulting KG looks very promising, it is not yet possible to provide a reliable measure. We are currently annotating true and false relations in a set of candidate sentences. This enables the evaluation of different prompts or different LLMs, as well as different settings for \lstinline|minRef|.

\section{Conclusion} \label{sec:conclusion}

In this paper, we have shown how to create a KG, which is linked to ChEBI, using the same vocabulary and extending it with knowledge from chemistry research papers. We see several applications for our approach:

Our KG can assist in extending the ChEBI ontology by suggesting chemical entities and roles that are not part of it.\footnote{For enhanced visibility, the corresponding namespace "CEAR" in Table~\ref{tab:cear_chems_and_roles} has been highlighted in bold.} Table~\ref{tab:cear_chems_and_roles} shows the top 10 most frequent relations with chemical entities and chemical roles not annotated in ChEBI. For example, PBS (phosphate-buffered saline) was correctly identified as a buffer 249 times in our set of 8,000 research papers. All text locations (the research paper, the page, and the character position of the relevant sentence) are available and can be used for reference. Future versions of CEAR will incorporate them using RDF-star. Extending the scope to larger collections of chemistry research papers can amplify the number of results for chemical entities and relations that are not annotated in ChEBI, thereby enhancing the usefulness of the KG.

\begin{table*}
    \centering
    \caption{Most frequent relations with chemical entities (top) and roles (bottom) which are not part of ChEBI}
    \label{tab:cear_chems_and_roles}
    \begin{tabular}{llllc}
        \toprule
        \textbf{source} & \textbf{chemical entity} & \textbf{source} & \textbf{chemical role} & \textbf{count} \\
        \midrule
        \textbf{CEAR} & PBS & ChEBI & buffer & 249 \\
        \textbf{CEAR} & CH2Cl2 & ChEBI & buffer & 117 \\
        \textbf{CEAR} & metal & ChEBI & catalyst & 76 \\
        \textbf{CEAR} & ACN & ChEBI & solvent & 62 \\
        \textbf{CEAR} & Tris-HCl & ChEBI & buffer & 45 \\
        \textbf{CEAR} & organolithium & ChEBI & reagent & 32 \\
        \textbf{CEAR} & terpyridine & ChEBI & ligand & 31 \\
        \textbf{CEAR} & Et2O & ChEBI & solvent & 31 \\
        \textbf{CEAR} & CH2Cl2 & ChEBI & reagent & 28 \\
        \textbf{CEAR} & metal & ChEBI & reagent & 26 \\
        \midrule
        ChEBI & hydrogen atom & \textbf{CEAR} & fuel & 33 \\
        ChEBI & ammonia & \textbf{CEAR} & fuel & 24 \\
        ChEBI & carbon dioxide & \textbf{CEAR} & feedstock & 19 \\
        ChEBI & methanol & \textbf{CEAR} & fuel & 16 \\
        ChEBI & hydrocarbon & \textbf{CEAR} & fuels & 15 \\
        ChEBI & ethanol & \textbf{CEAR} & fuel & 15 \\
        ChEBI & dihydrogen & \textbf{CEAR} & fuel & 15 \\
        ChEBI & methane & \textbf{CEAR} & fuel & 11 \\
        \textbf{CEAR} & gasoline & \textbf{CEAR} & fuel & 9 \\
        ChEBI & CCCP & \textbf{CEAR} & protonophore & 9 \\
        \bottomrule
    \end{tabular}
\end{table*}

Furthermore, we are developing exploration utilities for working with chemistry research papers. By detecting chemical entities and their roles, we can, for example, highlight them in the papers and direct users to ChEBI or PubChem for additional information.

\section{Acknowledgment}
This work was supported by the Research Initiative
"SmartProSys: Intelligent Process Systems for the Sustainable Production
of Chemicals" funded by the Ministry for Science, Energy, Climate
Protection and the Environment of the State of Saxony-Anhalt.

\bibliography{sps-knowledge-graph-creation}

\appendix

\section{Online Resources}

Our source code is available at:
\href{https://github.com/stlanger/cear}{https://github.com/stlanger/cear} \\
\\
The Turtle representation of the KG (using a \lstinline|minRef| hyperparameter of 2) is available at:
\href{https://wwwiti.cs.uni-magdeburg.de/iti_dke/cear/cear.ttl}{https://wwwiti.cs.uni-magdeburg.de/iti\_dke/cear/cear.ttl} \\
\\
Other versions with different settings for \lstinline|minRef| can be viewed and downloaded at:\\
\href{https://wwwiti.cs.uni-magdeburg.de/iti_dke/cear/cear.ttl}{https://wwwiti.cs.uni-magdeburg.de/iti\_dke/cear/}

\end{document}